\documentclass[times,referee,twocolumn,final,authoryear]{elsarticle}

\usepackage{ycviu}
\usepackage{framed,multirow}

\usepackage{amssymb}
\usepackage{latexsym}

\usepackage{url}
\usepackage{xcolor}
\usepackage{graphicx}

\usepackage{mathtools, cuted}
\usepackage{multirow}
\usepackage{booktabs}
\definecolor{newcolor}{rgb}{.8,.349,.1}

\journal{Computer Vision and Image Understanding}

\begin{document}

\thispagestyle{empty}

\clearpage

\ifpreprint
  \setcounter{page}{1}
\else
  \setcounter{page}{1}
\fi

\begin{frontmatter}

\title{Geometric Pose Affordance: Monocular 3D Human Pose Estimation with Scene Constraints}

\author[1]{Zhe \snm{Wang}} 
\author[1]{Liyan \snm{Chen}}
\author[1]{Shaurya  \snm{Rathore}}
\author[1]{Daeyun  \snm{Shin}}
\author[1]{Charless  \snm{Fowlkes}}

\address[1]{Department of Computer Science, University of California, Irvine, CA 92617, USA}

\received{1 May 2013}
\finalform{10 May 2013}
\accepted{13 May 2013}
\availableonline{15 May 2013}
\communicated{S. Sarkar}

\begin{abstract}
  Accurate estimation of 3D human pose from a single image remains a challenging task
despite many recent advances. In this paper, we explore the hypothesis that
strong prior information about scene geometry can be used to improve pose
estimation accuracy. To tackle this question empirically, we have assembled a
novel {\em Geometric Pose Affordance} dataset, consisting of multi-view
imagery of people interacting with a variety of rich 3D environments.  We
utilized a commercial motion capture system to collect gold-standard estimates
of pose and construct accurate geometric 3D models of the scene geometry.

To inject prior knowledge of scene constraints into existing frameworks for
pose estimation from images, we introduce a view-based representation of
scene geometry, a {\em multi-layer depth map}, which employs multi-hit ray
tracing to concisely encode multiple surface entry and exit points along each
camera view ray direction. We propose two different mechanisms for
integrating multi-layer depth information into pose estimation: 
input as encoded ray features used in lifting 2D pose to full 3D, and secondly
as a differentiable loss that encourages learned models to favor geometrically
consistent pose estimates. We show experimentally that these techniques can
improve the accuracy of 3D pose estimates, particularly in the
presence of occlusion and complex scene geometry.
\end{abstract}


\end{frontmatter}



\section{Introduction}

Accurate estimation of human pose in 3D from image data would enable a wide
range of interesting applications in emerging fields such as virtual and
augmented reality, humanoid robotics, workplace safety, and monitoring mobility and fall
prevention in aging populations. Interestingly, many such applications are set
in relatively controlled environments (e.g., the home) where large parts of the
scene geometry are relatively static (e.g., walls, doors, heavy furniture). We
are interested in the following question, {\em ``Can strong knowledge of scene
geometry improve our estimates of human pose from images?''}. 

\begin{figure*}[t]
\begin{center}
   \includegraphics[width=1\linewidth]{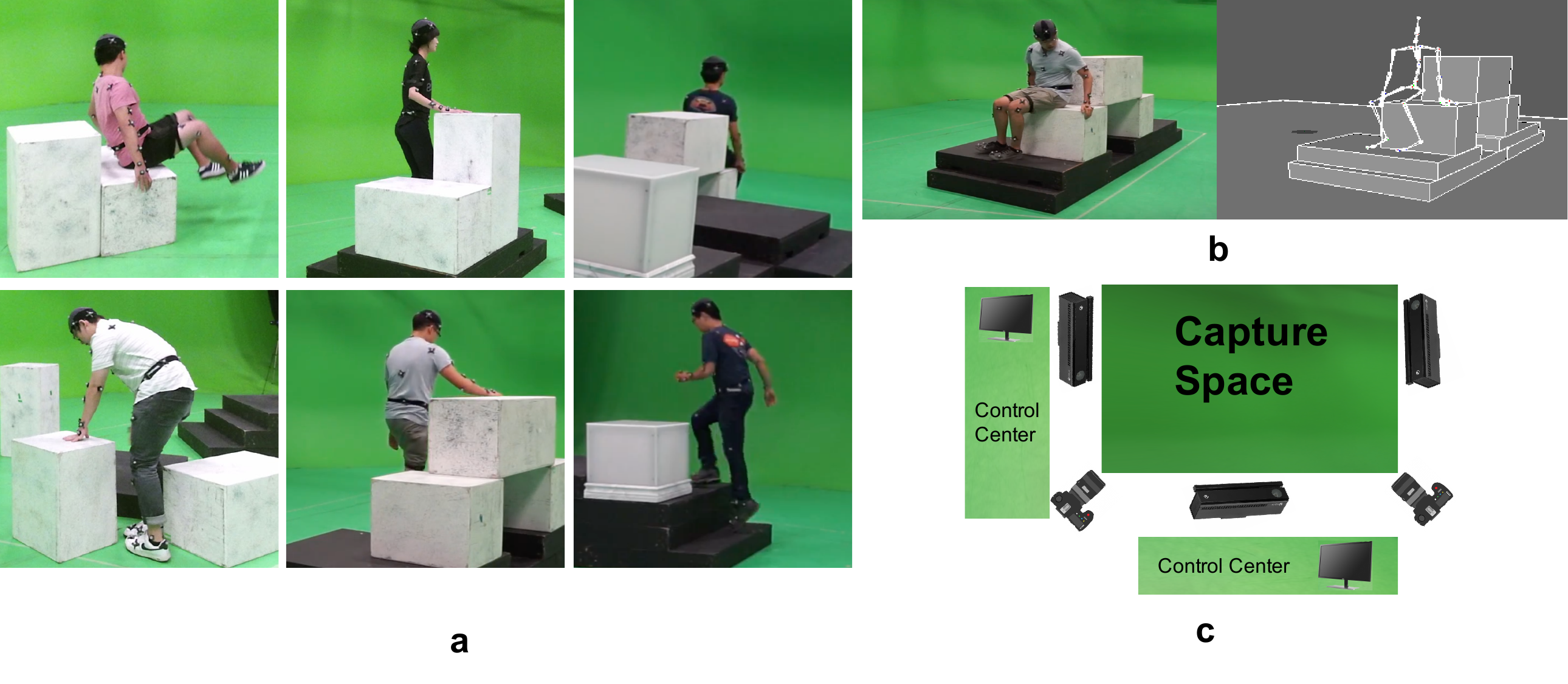}
\end{center}
   \caption{\textbf{a}: Samples from our data set featuring scene constrained poses:
   stepping on the stairs, sitting on the tables and touching boxes. 
   \textbf{b}: Sample frame of a human interacting with scene geometry, and visualization
   of the corresponding 3D scene mesh with captured human pose.
   \textbf{c}: Motion capture setup. We simultaneously captured 3 RGBD and 2 RGB video 
   streams and ground-truth 3D pose from a VICON marker-based mocap system. Cameras
   are calibrated with respect to a 3D mesh model of scene geometry.
   }
\label{fig:capture}
\end{figure*}

Consider the images in Fig. \ref{fig:capture} a. Intuitively, if we know the 3D
locations of surfaces in the scene, this should constrain our estimates of pose.
Hands and feet should not interpenetrate scene surfaces, and if we see someone
sitting on a surface of known height we should have a good estimate of where
their hips are even if large parts of the body are occluded. This general
notion of scene affordance~\footnote {``The meaning or value of a thing consists
of what it affords.'' -JJ Gibson (1979)} has been explored as a tool for
understanding functional and geometric properties of a scene
\citep{Guptaafford,Fouhey12,Wang_affordanceCVPR2017,xueting}.  However, the focus of
such work has largely been on using estimated human pose to infer scene
geometry and function.

Surprisingly, there has been little demonstration of how scene knowledge can
constrain pose estimation. Traditional 3D pose estimation models have explored 
kinematic and dynamic constraints which are scene agnostic and have been tested
on datasets of people freely performing actions in large empty spaces.  
{\em We posit one reason
that scene constraints have not been utilized is lack of large-scale datasets
of annotated 3D pose in rich environments.} Methods have been developed on datasets like
Human3.6M \citep{h36m_pami} and MPI-INF-3DHP \citep{mono_3dhp2017}, which lack
diverse scene geometry (at most one chair or sofa) and are generally free from scene 
occlusion.  Recent efforts have allowed for more precise 3D pose capture for in-the-wild 
environments \citep{inthewildeccv2018} but lack ground-truth scene geometry, or provide
scene geometry but lack extensive ground-truth pose estimates \citep{mpii_scene}.

Instead of tackling human pose estimation in isolation, we argue that systems
should take into account available information about constraints imposed by
complex environments. A complete solution must ultimately tackle two problems:
(i) estimating the geometry and free space of the environment (even when much
of that free space is occluded from view), (ii) integrating this information
into pose estimation process. Tools for building 3D models of static
environments are well developed and estimation of novel scene geometry from
single-view imagery has also shown rapid progress. Thus, we focus on the second
aspect under the assumption that high-quality geometric information is available
as an input to the pose estimation pipeline.

The question of how to represent geometry and incorporate the 
constraints it imposes with current learning-based approaches to modeling human
pose is an open problem.  There are several candidates for representing scene
geometry: voxel representations of occupancy \citep{volumetric} are
straightforward but demand significant memory and computation to achieve
reasonable resolution; Point cloud \citep{pointcloud} representations provide
more compact representations of surfaces by sampling but lack topological
information about which locations in a scene constitute free space. Instead, we
propose to utilize {\em multi-layer depth maps}~\citep{mdp} which provide a compact 
and nearly complete representation of scene geometry that can be readily
queried to verify pose-scene consistency.

We develop and evaluate several approaches to utilize information contained in the 
multi-layer depth map representation. Since multi-layer depth is a view-centered 
representation of geometry, it can be readily incorporated as an additional input 
feature channel. We leverage estimates of 2D pose either as a heatmap or regressed 
coordinate and query the multi-layer depth map directly to extract features encoding 
local constraints on the z-coordinates of joints that can be used to predict
geometry-aware 3D joint locations.  Additionally, we introduce a differentiable
loss that encourages a model trained with such features to respect hard
constraints imposed by scene geometry.  We perform an extensive evaluation of
our multi-layer depth map models on a range of scenes of varying complexity and
occlusion.  We provide both qualitative and quantitative evaluation on real
data demonstrating that these mechanisms for incorporating geometric constraints
improves upon scene-agnostic state-of-the-art methods for 3D pose estimation. 

To summarize our main contributions:
1. We collect and curate a unique, large-scale 3D human pose estimation
dataset with rich ground-truth scene geometry and a wide variety of pose-scene
interactions (see e.g. Fig. \ref{fig:capture})
2. We propose a novel representation of scene geometry constraints:
multi-layer depth map, and explore multiple ways to incorporate geometric
constraints into contemporary learning-based methods for predicting 3D human
pose.
3. We experimentally demonstrate the effectiveness of integrating
geometric constraints relative to two state-of-the-art scene-agnostic
pose estimation methods.

\begin{table*}[!ht]
\small
\begin{center}
\begin{tabular}{|l|c|c|c|}
\hline
Dataset & Frames & Scenes & Characteristics \\
\hline
HumanEva (2010)  & 80k & ground plane & marker-based pose and video \\
Human36M (2014) & 3.6M & chairs & marker-based, human body scans \\
MPI-INF-3DHP (2017) & 3k & chairs, sofa & marker-less, indoor and outdoor backgrounds\\
TotalCapture (2017) & 1.9M & ground plane  & marker-based pose, IMU and video \\
Surreal (2017) & 6M & ground plane  & synthetic renderings of CMU Mocap data \\
Ski-Pose (2018) & 10k & ski slope  & marker-less using multi-view 2D annotation\\
3DPW (2018) & 51k & in the wild  & IMU-based capture with mobile camera\\
\textbf{GPA (2019)} & 0.7M & boxes, chairs, stairs  & scene interaction, occlusion, geometry ground-truth\\
\hline
\end{tabular}
\end{center}
\caption{Comparison of existing datasets commonly used for training and evaluating 3D human pose estimation methods. Previous datasets have primarily focused on capturing a diverse range human motions, actions, abd subjects using optical markers and/or IMUs to establish ground-truth pose. Our dataset focuses on interactions between humans and static scene geometry and includes both ground-truth 3D pose and a complete description of the scene geometry.}
\label{table:datasets}
\end{table*}

\section{Related Work}

\paragraph{Motion capture for ground-truth 3D pose}
The work of \cite{humaneva} introduced one of the first large-scale 3D human pose estimation
datasets with synchronized images and ground-truth 3D keypoint locations.
\cite{h36m_pami} scaled their dataset up to 3.6 million images covering a range of
subjects and actions along with depth images and 3D body scans of the human subjects.
To overcome the limitations of marker-based data collection such as constrained clothing and 
capture environment, several marker-less approaches have also been used.  \cite{taotal} utilize 
an indoor "panoptic studio" to capture poses from 10 calibrated RGBD cameras.  \cite{mono_3dhp2017} 
utilized multi-view marker-less capture to collect pose data for subjects wearing a variety of 
clothing against both indoor and outdoor backgrounds.  \cite{cvpr18multiviewepfl} utilized calibrated PTZ 
cameras and human annotators to triangulate joint locations skiers over a large area of a ski-slope.
\cite{zhou2018drocap} also explores motion capture both indoor and outdoor using a Drone. 
Synchronized inertial measurement sensor (IMU) data can be used to further enhance marker-less 
capture. \cite{Trumble:BMVC:2017} develop an approach to fusing inertial measurement sensors with
multi-view recording in a studio environment. \cite{mono_3dhp2017} use an IMU-based system along 
with a single synchronized mobile camera video stream to capture 3D human pose "in the wild".

These data collection efforts have largely focused on covering a diverse range of poses and 
actions, but actions take place in simple environments (i.e., an empty room) which minimize
occlusion and impose very few geometric affordance constraints on human pose. Recent "in the wild" 
markerless capture data such as \cite{mono_3dhp2017} encompass much richer environments, but the 
scene geometry is unknown.  In contrast, our dataset provides gold-standard, marker-based 3D pose 
of subjects in richer environments with ground-truth scene geometry, offering a controlled 
test-bed for research in 3D human pose estimation with rich geometric affordance.
\cite{GRAB:2020} collects a dataset for grasping, with the markers placed both on hands and on bodies to capture whole-body pose during grasping and object manipulation. This is complementary to our dataset as it provides object geometry and 
grasping contacts while our dataset samples whole-body affordance. We 
provide a summary comparison of recent 3D human pose estimation datasets in Table \ref{table:datasets}.

\paragraph{Modeling scene affordances} 
The term ``affordance'' was coined by J Gibson
\citep{gibson} to capture the notion that the meaning and relevance of many
objects in the environment are largely defined in relation to the ways in which
an individual can functionally interact with them. For computer vision, this
suggests scenarios in which the natural labels for some types of visual content
may not be semantic categories or geometric data but rather functional labels, i.e., 
which human interactions they afford.  \cite{Guptaafford} present a human-centric 
paradigm for scene understanding by modeling physical human-scene interactions.
\cite{Fouhey12} rely on pose estimation methods to extract functional and
geometric constraints about the scene and use those constraints to improve
estimates of 3D scene geometry.  \cite{Wang_affordanceCVPR2017} collects a
large-scale dataset of images from sitcoms which contains multiple images 
of the same scene with and without humans present. Leveraging state-of-the-art 
pose estimation and generative model to infer what kind of poses 
each sitcom scene affords.  \cite{xueting} build a  fully  automatic  3D  pose  
synthesizer to predict semantically plausible and physically feasible human poses 
within a given scene.  \cite{iMapper} applies an energy-based model on synthetic 
videos to improve both scene and human motion mapping. 
\cite{caoHMP2020} construct a synthetic dataset utilizing a game engine.  They first sample multiple human motion goals based on a single scene image and 2D pose histories, plan 3D human paths towards each goal, and finally predict 3D human pose sequences following each path.
Rather than labeling image 
content based on observed poses, our approach is focused on estimating 
scene affordance directly from physical principles and geometric data, and then
subsequently leveraging affordance to constrain estimates of human pose and 
interactions with the scene.

Our work is also closely related to earlier work on scene context for object
detection.  \cite{Hoiem05,Hoiem06} used estimates of ground-plane geometry to reason
about location and scales of objects in an image. More recent work such as
\cite{Wang_2015_CVPR,diaz2016lifting,Matzen13} use more extensive 3D models of scenes 
as context to improve object detection performance.  Geometric context for human pose 
estimation differs from generic object detection in that humans are highly articulated. 
This makes incorporating such constraints more complicated as the resulting predictions 
should simultaneously satisfy both scene-geometric and kinematic constraints. 

\paragraph{Constraints in 3D human pose estimation} Estimating 3D human pose
from monocular image or video is an ill-posed problem that can benefit from
prior constraints. Recent examples include \cite{rnnpose} who model kinematics, 
symmetry and motor control using an RNN when predicting 3D human joints directly
from 2D key points. \cite{adversialpose} propose an  adversarial network as an 
anthropometric regularizer.  \cite{graphicalmodel,zhou2018monocap} construct a 
graphical model encoding priors to fit 3D pose reconstruction.
\cite{LCRnet++,posematching} first build a large set of valid 3D human poses and 
treat estimation as a matching or classification problem.
\cite{jointconstraint,geometryaware} explore joint constraints in 3D and
geometric consistency from multi-view images.  \cite{Zhou_2017_ICCV} improve
joint estimation by adding bone-length ratio constraints.

To our knowledge, there is relatively little work on utilizing scene constraints
for 3D human pose. \cite{groundconstraint} utilize an energy-based optimization model for
pose refinement which penalizes ankle joint estimates that are far above or below 
an estimated ground-plane. The recent work of \cite{mpii_scene} introduces
scene geometry penetration and contact constraints in an energy-based framework for 
fitting parameters of a kinematic body model to estimate pose. In our work, we 
explore a complementary approach which uses CNN-based regression
models that are trained to directly predict valid pose estimates given image and 
scene geometry as input.

\section{Geometric Pose Affordance Dataset (GPA)}

To collect a rich dataset for studying interaction of scene geometry and human
pose, we designed a set of action scripts performed by 13 subjects, each of
which takes place in one of 6 scene arrangements.  In this section, we describe
the dataset components and the collection process.  

\begin{figure*}[!bht]
\begin{center}
   \includegraphics[width=0.9\linewidth]{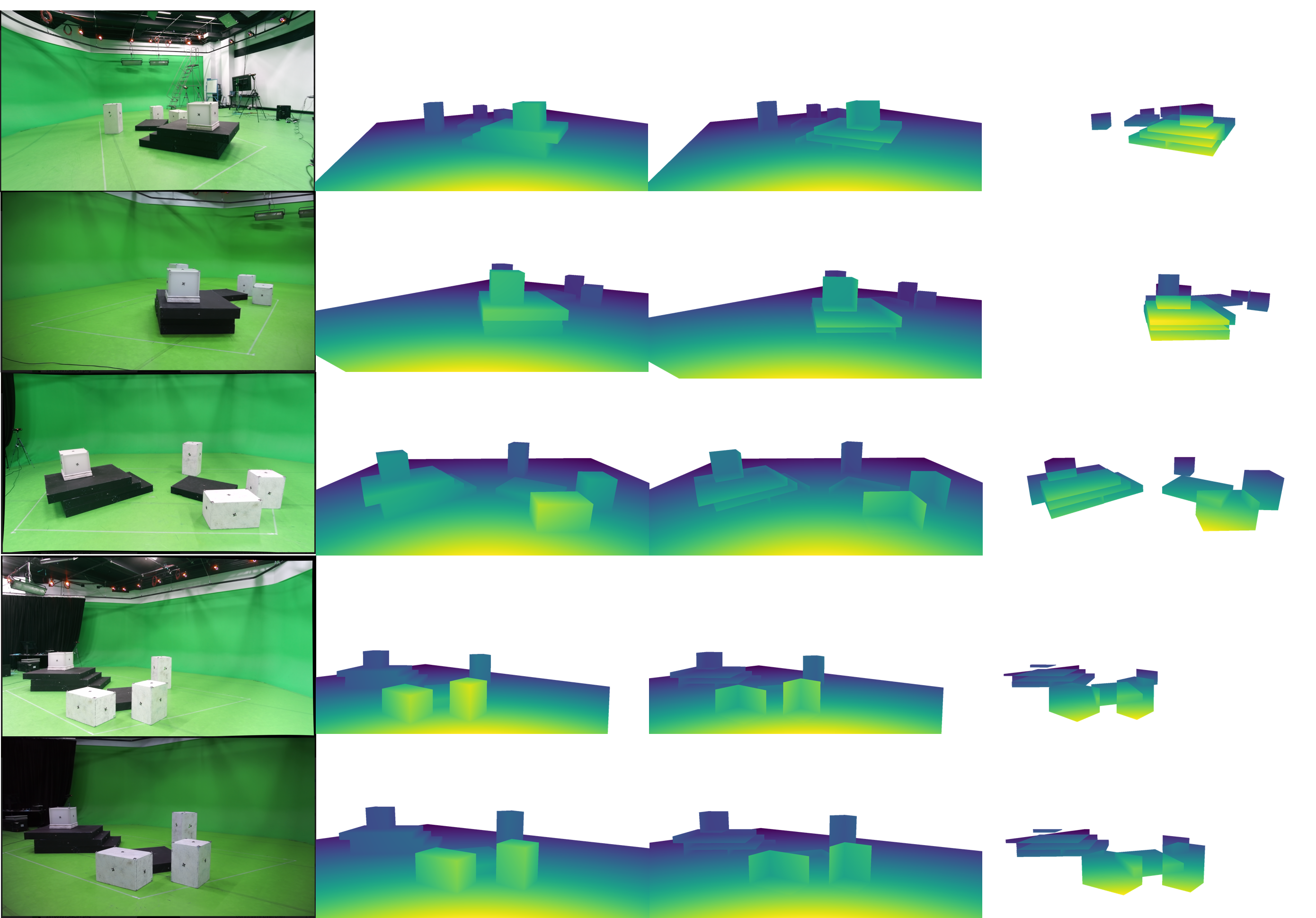}
\end{center}
   \caption{The 5 camera views from the same scene with the
   first 3 layers of corresponding multi-layer depth map (for visualization
   clarity, we plot inverse depth). 2nd column corresponds to a traditional
   depth map, recording the depth of the first visible surface in the scene from
   the camera viewpoint of 1st column. 3rd column is when the multi-hit ray
   leaves the first layer of objects (e.g. the backside of the boxes). 4th column is 
   when the multi-hit ray hits another object.
   }
\label{fig:camera01234_mds}
\end{figure*}

\subsection{Human Poses and Subjects}
We designed three action scripts that place emphasis on semantic actions, 
mechanical dynamics of skeletons, and pose-scene interactions.
We refer to them as \textit{Action, Motion, and Interaction Sets} respectively.
The semantic actions of \textit{Action Set} are constructed from a subset
of Human3.6M~\citep{h36m_pami}, namely, \textit{Direction, Discussion, Writing,
Greeting, Phoning, Photo, Posing} and \textit{Walk Dog} to provide a connection
for comparisons between our dataset and the de facto standard benchmark.
\textit{Motion Set} includes poses with more dynamic range of motion, such as
running, side-to-side jumping, rotating, jumping over obstacles, and improvised poses
from subjects.
\textit{Interaction Set} mainly consists of close interactions between body 
parts and surfaces in the scene to support modeling geometric affordance in 3D.
There are three main poses in this group: \textit{Sitting, Touching, Standing on},
corresponding to typical affordance relations \textit{Sittable, Walkable, Reachable}
\citep{Fouhey12,Guptaafford}.
The 13 subjects included 9 males and 4 female with roughly the same age and
medium variations in heights approximately from 155cm to 190cm, giving comparable
subject diversity to Human3.6M.

\subsection{Image Recording and Motion Capture} 
This motion capture studio layout is also illustrated in Fig. \ref{fig:capture} c.
We utilized two types of camera, RGBD and RGB, placed at 5 distinct locations in the 
capture studio. All 5 cameras have a steady 30fps frame rate but their time stamps 
are only partially synchronized, requiring additional post-processing described below.  
The color sensors of the 5 cameras have the same 1920x1080 resolution and the depth
sensor of the Kinect v2 cameras has a resolution at 640x480.  The motion capture system 
was a standard VICON 
system with 28 pre-calibrated cameras covering the capture space which are used to
estimate the 3D coordinates of IR-reflective tracking markers attached to the surface 
of subjects and objects.

\subsection{Scene Layouts}
Unlike previous efforts that focus primarily on human poses without other objects present
(e.g.~\citep{h36m_pami,mono_3dhp2017}), we introduced a variety of scene geometries
with arrangements of 9 cuboid boxes in the scene. The RGB images captured from
5 distinct viewpoints exhibit substantially more occlusion of subjects than existing 
datasets (as illustrated in Fig  \ref{fig:capture} and Fig \ref{fig:camera01234_mds})
and constrain the set of possible poses. We captured 1 or 2 subjects interacting with 
each scene and configured a total of 6 distinct scene geometries.

To record static scene geometry, we measured physical dimension of all the objects
(cuboids) as well as scanning the scene with a mobile Kinect sensor.  We utilized 
additional motion-capture markers attached to the corners and center face of each 
object surface so that we could easily align geometric models of the cuboids with the 
global coordinate system of the motion capture system. We also use the location of 
these markers, when visible in the RGB capture cameras, in order to estimate
extrinsic camera parameters in the same global coordinate system.  This allows
us to quickly create geometric models of the scene which are well aligned to
all calibrated camera views and the motion capture data.

\begin{figure*}
\begin{center}
  
   \includegraphics[width=0.9\linewidth]{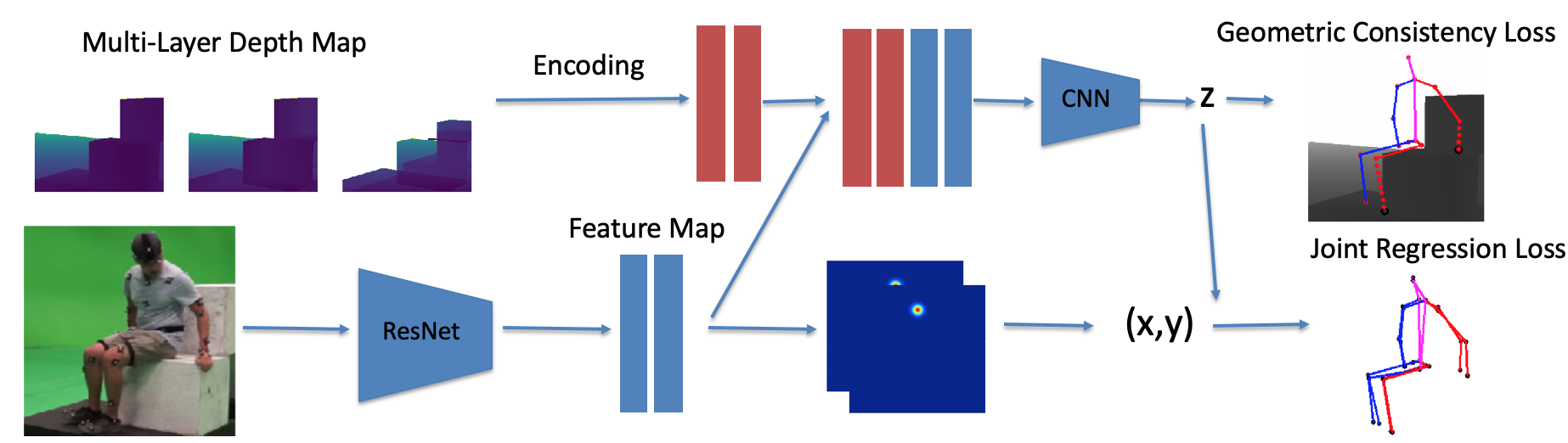}
\end{center}
   \caption{Overview of model architecture: we use ResNet-50 as our backbone to
   extract features from a human centered cropped image.  The feature map is
   used to predict 2D joint location heatmaps and is also concatenated with
   encoded multi-layer depth map. The concatenated feature is used to regress
   the depth (z-coordinate) of each joint. The model is trained with a loss
   on joint location (joint regression loss) and scene affordance (geometric
   consistency loss). The 2d joint heatmaps are decoded to x,y joint locations 
   using an argmax. The geometric consistency loss is described in more detail in
   Fig \ref{fig:constraint} (a) and Section 4.2.} 
\label{fig:method}
\end{figure*}

\subsection{Scene Geometry Representation}
\begin{figure}
\begin{center}
   \includegraphics[width=0.9\linewidth]{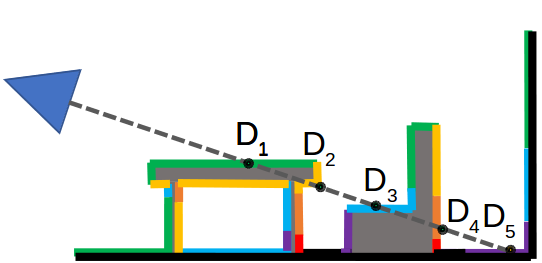}
\end{center}
   \caption{Illustration of multi-layer depth map. For each image pixel we record the 
   depth of all surface intersections along the view ray (e.g., $D_1, D_2, D_3, D_4, D_5$ ).} 
\label{fig:multilayer}
\end{figure}

Mesh models of each scene were initially constructed in global coordinates using
modeling software (Maya) with assistance from physical measurements and reflective 
markers attached to scene objects. To compactly represent the scene geometry from the 
perspective of a given camera viewpoint, we utilize a multi-layer depth map.
{\em Multi-layer depth maps} are defined as a map of camera ray entry and exit depths for all 
surfaces in a scene from a given camera viewpoint (illustrated in Fig~\ref{fig:multilayer}).
Unlike standard depth-maps which only encode the geometry of visible surfaces in a scene 
(sometimes referred to as 2.5D), multi-layer depth provides a nearly\footnote{Surfaces 
tangent to a camera view ray are not represented} complete, viewer-centered description of 
scene geometry which includes occluded surfaces.

The multi-layer depth representation can be computed from the scene mesh model by 
performing multi-hit ray tracing from a specified camera viewpoint. Specifically, the 
multi-hit ray tracing sends a ray from the camera center towards a point on the 
image plane that corresponds to the pixel at $(x, y)$ and outputs distance
values $\{t_1, t_2, t_3, ..., t_k\}$ where $k$ is the total number of
polygon intersections along the ray.  Given a unit ray direction $\mathbf{r}$ and
camera viewing direction $\mathbf{v}$, the depth value at layer $i$ is $D_i(x, y) =
t_i\mathbf{r}\cdot\mathbf{v}$ if $i <= k$ and $D_i(x, y) = \varnothing$ if $i >
k$. In our scenes, the number of multi-layer depth maps is set to 15 which suffices
to cover all scene surfaces in our dataset. We visualize 5 camera viewpoints together 
with first 3 layers of depth map in the same scene in Fig \ref{fig:camera01234_mds}.

\subsection{Data Processing Pipeline}
The whole data processing pipeline includes validating motion capture pose
estimates, camera calibration, joint temporal alignment of all data  sources,
and camera calibration.
Unlike previous marker-based mocap datasets which have few occlusions, many markers
attached to the human body are occluded in the scene during our capture
sessions due to scene geometry. We spent 4 months on pre-processing with help of 6 
annotators in total.  There are three stages of generating ground truth joints from 
recorded VICON sessions: \textbf{(a)} recognizing and labeling recorded markers in each
frame to 53 candidate labels which included three passes to minimize errors;
\textbf{(b)} applying selective temporal interpolation for missing markers
based on annotators' judgement. \textbf{(c)} removing clips with too few
tracked markers. After the annotation pipeline, we compiled recordings and
annotations into 61 sessions captured at 120fps by the VICON software. To
temporally align these compiled ground-truth pose streams to image capture
streams, we first had annotators to manually correspond 10-20 pose
frames to image frames. Then we estimated temporal scaling and offset 
parameters using RANSAC \cite{fischler1981random}, and regress all timestamps 
to a single global timeline. 

The RGB camera calibration was performed by having annotators mark 
corresponding image coordinates of visible markers (whose global 3D coordinates 
are known) and estimating extrinsic camera parameters from those correspondences. We performed 
visual inspection on all clips to check that the estimated camera parameters yield 
correct projections of 3D markers to their corresponding locations in the image.
With estimated camera distortion parameters, we correct the radial and lens distortions 
of the image so that they can be treated as projections from ideal pinhole cameras 
in later steps. Finally, the scene geometry model was rendered into multi-layer 
depth maps for each calibrated camera viewpoint. We performed visual inspection to 
verify that the depth edges in renderings were precisely aligned with object boundaries 
in the RGB images.

After temporal and geometric calibration, we generated a unified dataset by using
an adaptive sampling approach to select non-redundant frames. We consider frames 
with sufficiently different poses from adjacent ones as ``interesting''.  Here, the
measure of difference between two skeleton poses is defined as the 75th percentile
of L2 distances between corresponding joints (34 pairs per skeleton pair).
This allows us to retain frames where only a few body parts moved significantly
while being robust to inter-frame differences due to noise or missing markers.
With the measure of difference defined, we select the frames by choosing the 
change threshold as the 55th percentile, retaining 45\% of total frames from
the original sequences. This final dataset, which we call Geometric Pose Affordance 
(GPA) contains 304.9k images, each with corresponding ground-truth 3D pose and 
scene geometry\footnote{The dataset is available online: {\url{https://wangzheallen.github.io/GPA}}}. 

\subsection{Dataset Visualization and Statistics}

A video demonstrating the output of this pipeline is available online~\footnote 
{ Video Link: {\url{https://youtu.be/ZRnCBySt2fk}}}. The video shows the full 
frame and a crop with ground-truth joints/markers overlayed, for 10 sample clips 
from the 'Action' and 'Motion' sets.  The video also indicates various diagnostic 
metadata including the video and mocap time stamps, joint velocities, and number of 
valid markers (there are 53 markers and 34 joints for VICON system). Since we have 
an accurate model of the scene geometry, we can also automatically determine which 
joints and markers are occluded from the camera viewpoint.

\begin{figure}
\centering
\includegraphics[width=0.9\linewidth]{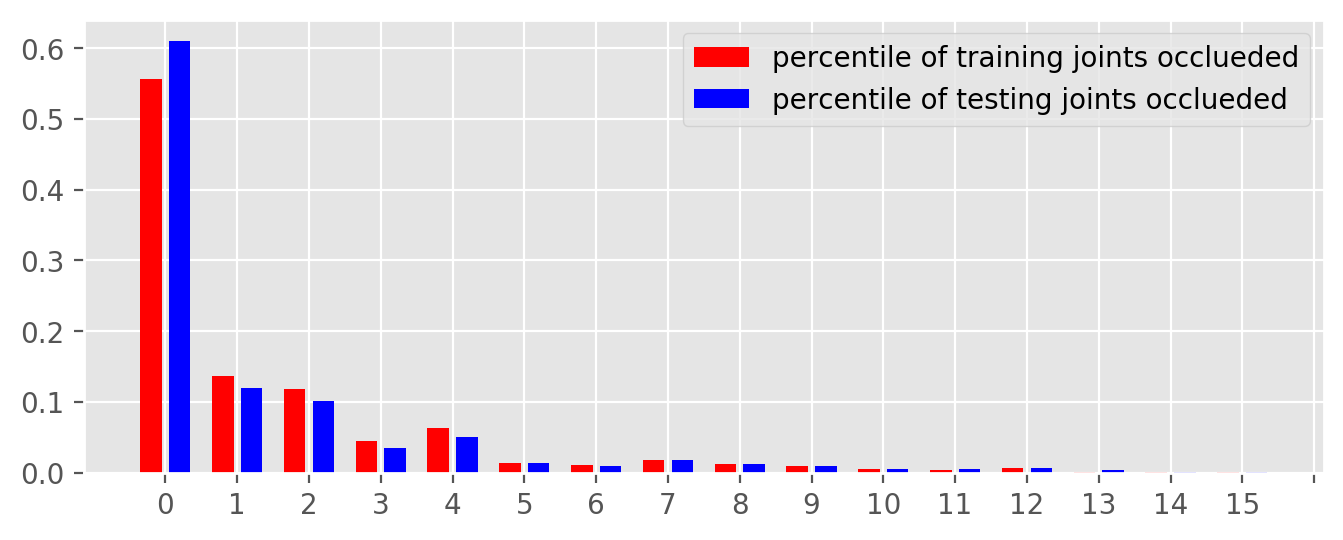}
\includegraphics[width=0.9\linewidth]{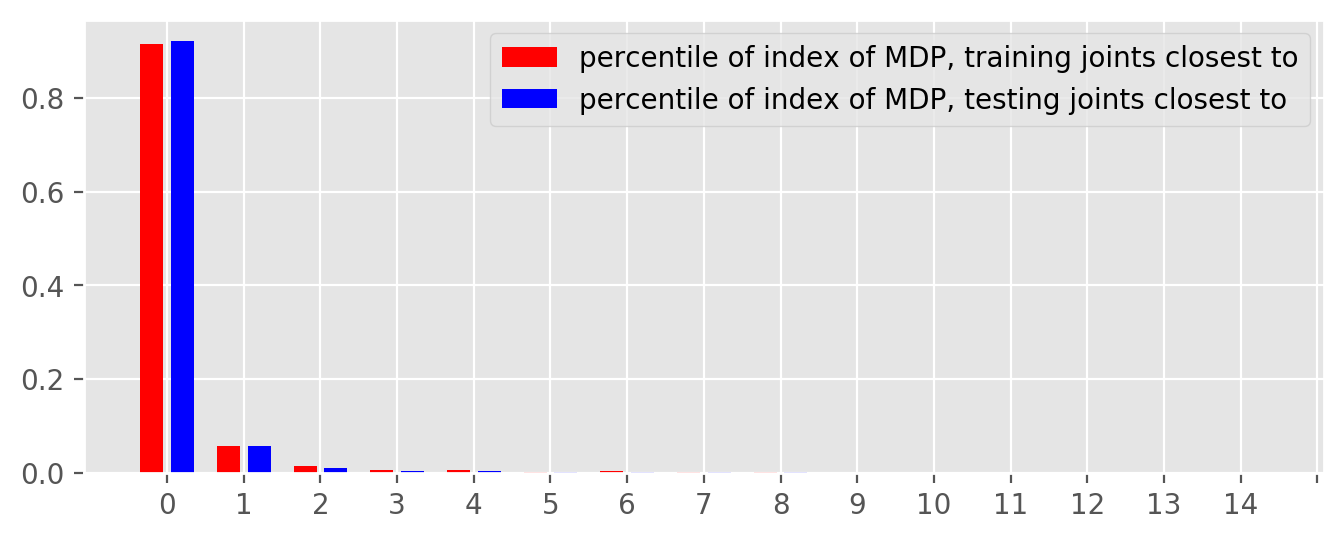}
\caption{ Top: Distribution of the number of joints occluded in training and
testing frames. Bottom: Distribution of the index of the depth layer closest 
to each pose. High index layers, which often correspond to hidden surfaces such
as the bottom side of platforms, seldom constrain pose.}
\label{fig:occlusion}
\end{figure}

Fig. \ref{fig:occlusion} summarizes statistics on the number of occluded joints 
as well as the distribution of which multi-depth layer is closest to a joint. 
While the complete scene geometry requires 15 depth layers, as the figure shows
only the first 5 layers are involved in 90\% of the interaction between body 
joints and scene geometry. The remaining layers often represent surfaces which are 
inaccessible (e.g., bottoms of cuboids).

\section{Geometry-aware Pose Estimation}

We now introduce two approaches for incorporating geometric affordance in
CNN-based pose regression, building on the baseline architecture of
\cite{Zhou_2017_ICCV}.  Given an image $I$ of a human subject, we aim to
estimate the 3D human pose represented by a set of 3D joint coordinates of the
human skeleton, $P \in \mathbb{R}^{J \times 3}$ where $J$ is the number of
joints.  We follow the convention of representing each 3D coordinate in
the local camera coordinate system associated with $I$. The first two
coordinates are given by image pixel coordinates and the third coordinate is
the joint depth in metric coordinates (e.g., millimeters) relative to the
depth of a specified root joint. We use $P_{XY}$ and $P_{Z}$ respectively as 
short-hand notations for the components of $P$.

\subsection{Pose Estimation Baseline Model}
We adopt one popular ResNet-based network described by \cite{xiao2018simple} 
as our 2D pose estimation module. The network output is a set of low-resolution
heat-maps $\hat{S} \in \mathbb{R}^{64 \times 64 \times J}$, where each map
${\hat S}[:,:,j]$ can be interpreted as a probability distribution over the j-th joint
location.  At test time, the 2D prediction ${\hat P}_{XY}$ is given by the most
probable ($\arg\max$) locations in $S$.  This heat-map representation is
convenient as it can be easily combined (e.g., concatenated) with the other
spatial feature maps. To train this module, we utilize squared error loss 
\begin{equation}
{\ell_{2D}({\hat S}|P)} = \| {\hat S} - G(P_{XY}) \|^2
  \label{eqn:l2d}
\end{equation}
where $G(\cdot)$ is a target distribution created from ground-truth $P$ by placing
a Gaussian with $\sigma=3$ at each joint location. 

To predict the depth of each joint, we follow the approach of
\cite{Zhou_2017_ICCV}, which combines the 2D joint heatmap and the
intermediate feature representations in the 2D pose module as input to a joint
depth regression module (denoted {\bf ResNet} in the experiments). These shared 
visual features provide additional cues for recovering full 3D pose. We train 
with a smooth $\ell_1$ loss
\cite{fasterRCNN} given by:

\begin{equation}
  \ell_{1s}({\hat P}|P) =
    \begin{cases}
      \text{$\frac{1}{2} \| {\hat P}_Z- P_Z \|^2$ }  & \| {\hat P}_Z- P_Z \|\leq 1\\
      \text{$\| {\hat P}_Z - P_Z \| - \frac{1}{2}$} &  \text{o.w.}\\
    \end{cases}       
  \label{eqn:l1}
\end{equation}

\paragraph{Alternate baseline:} We also evaluated two alternative 
baseline architectures. First, we used the model of \cite{simple} which detects 2D joint 
locations and then trains a multi-layer perceptron to regress the 3D coordinates $P$ from 
the vector of 2D coordinates $P_{XY}$. We denote this simple lifting model as {\bf SIM} 
in the experiments.  To detect the 2D locations we utilized the ResNet model of \cite{xiao2018simple} 
and also considered an upper-bound based on lifting the ground-truth 2D joint locations to 3D.
Second, we trained the {\bf PoseNet} model proposed in \cite{rootnet} which uses integral 
regression \citep{integral} in order to regress pose from the heat map directly.

\subsection{Geometric Consistency Loss and Encoding}

To inject knowledge of scene geometry we consider two approaches, {\em 
geometric consistency loss} which incorporates scene geometry during training, 
and {\em geometric encoding} which assumes scene geometry is also available as 
an input feature at test time.

\begin{figure}[t]
\begin{center}
   \includegraphics[width=1\linewidth]{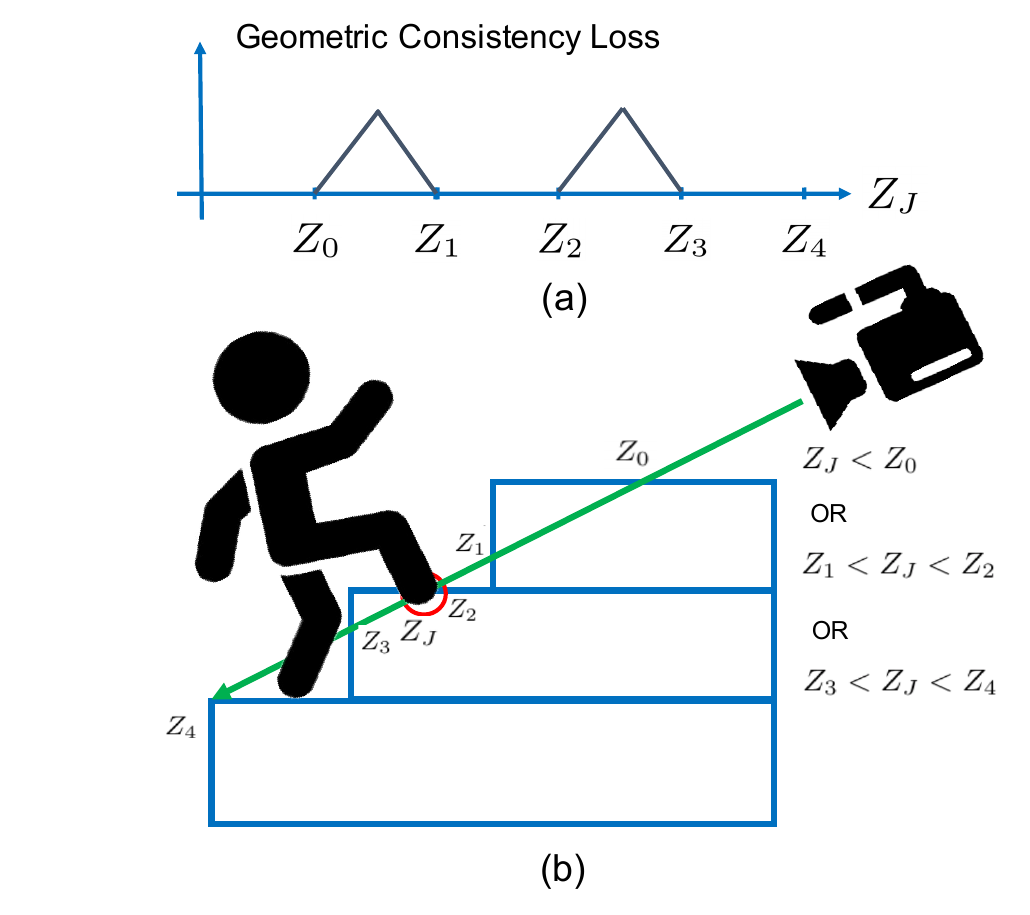}
\end{center}
   \caption{(a) is the illustration of the geometry consistency loss as a function of
   depth along a specific camera ray corresponding to a predicted 2D joint
   location.  In (b) the green line indicates the ray corresponding to the 2D
   location of the right foot. Our multi-depth encoding of the scene geometry
   stores the depth to each surface intersection along this ray (i.e., the
   depth values $Z_0, Z_1, Z_2, Z_3, Z_4$). Valid poses must satisfy the constraint
   that the joint depth falls in one of the intervals: $ Z_J < Z_0 $  or $ Z_1 <
   Z_J < Z_2 $ or $ Z_3 < Z_J < Z_4 $. The geometric consistency loss pushes the
   prediction $Z_J$ towards the closest valid configuration along the ray, $Z_J = Z_2$ .}
\label{fig:constraint}
\end{figure}

\paragraph{Geometric consistency loss:}
We design a geometric consistency loss (GCL) that specifically penalizes errors in 
pose estimation which violate scene geometry constraints. The intuition is illustrated
in Fig. \ref{fig:constraint}. For a joint at 2D location $(x,y)$, the estimated depth
$z$ should lie within one of a disjoint set of intervals defined by the multi-depth
values at that location.

To penalize a joint prediction $P^j=(x,y,z)$ that falls inside a region bounded by
front-back surfaces with depths $D_i(x,y)$ and $D_{i+1}(x,y)$ we define a loss that
increases linearly with the penetration distance inside the surface:
\begin{equation}
\begin{aligned}
  \ell_{G(i)}({\hat P}^j|D)  =  min( 
  &max(0, {\hat P}^j_Z - D_i({\hat P}^j_{XY})),\\
  &max(0, D_{i+1}({\hat P}^j_{XY}) - {\hat P}^j_Z ))
  \end{aligned}
\end{equation}
Our complete geometric consistency loss penalizes predictions which place 
any joint inside the occupied scene geometry
\begin{equation}
  \ell_{G}({\hat P}|D)  =  \sum_j \max_{i \in \{0,2,4,\ldots\}} \ell_{G(i)}({\hat P}^j|D)
  \label{eqn:lg}
\end{equation}
Assuming $\{D_i\}$ is piece-wise smooth, this loss is differentiable almost
everywhere and hence amenable to optimization with stochastic gradient descent. 
The gradient of the loss ``pushes'' joint location predictions for a given example 
to the surface of occupied volumes in the scene.

\paragraph{Encoding local scene geometry:}
When scene geometry is available at test time (e.g., fixed cameras pointed at a
known scene), it is reasonable to provide the model with an encoding of the
scene geometry as input. Our view-centered multi-depth representation of scene 
geometry can be naturally included as an additional feature channel in a CNN
since it is the same dimensions as the input image. 
We considered two different encodings of multi-layer depth. (1) We crop the
multi-layer depth map to the input frame, re-sample to the same resolution as the 2D
heatmap using nearest-neighbor interpolation, and offset by the depth of the
skeleton root joint.  (2) Alternately, we consider a volumetric encoding of the 
scene geometry by sampling 64 depths centered around the root joint using a range 
based on the largest residual depth between the root and any other joint seen during 
training (approx. $+/- 1m$). For each $(x,y)$ location and depth, we evaluate the 
geometric consistency loss $\ell_G$ at that point. This resulting encoding is
of size $H \times W \times 64$ and encodes the local volume occupancy around
the pose estimate. 

For the joint depth regression-based models ({\bf ResNet-*}) 
we simply concatenated the 
encoded multi-depth as additional feature channels.  For the lifting-based models
({\bf SIM-*}), we query the multi-depth values at the predicted 2D joint locations 
and use the results as additional inputs to the lifting network.

In our experiments we found that the simple and memory efficient multi-layer depth 
encoding (1) performed the same or better than volumetric encoding with ground-truth 
root joint offset. However, the volumetric encoding (2) was more robust when there was 
noise in the root joint depth estimate.

\subsection{Overall Training}

Combining the losses in Eq.  \ref{eqn:l2d}, \ref{eqn:l1}, and
\ref{eqn:lg}, the total loss for each training example is 
\begin{equation*}
\begin{aligned} 
 \ell({\hat P},{\hat S} | P,D) = \ell_{2D}({\hat S}|P) + \ell_{1s}({\hat P}|P) + \ell_{G}({\hat P}|P,D)
\end{aligned}
\end{equation*}
We follow \cite{Zhou_2017_ICCV} and adopt a stage-wise training approach: Stage
1 initializes the 2D pose module using 2D annotated images (i.e., MPII dataset);
Stage 2 trains the 3D pose estimation module, jointly optimizing the depth regression
module as well as the 2D pose estimation module; Stage 3 of training adds the
geometry-aware components (encoding input, geometric consistency loss) to the 
modules trained in stage 2.

\begin{table}[t]
\begin{center}
\small
\begin{tabular}{c|c}
\hline
\textbf{Set} & \textbf{Number of Images} \\ 
\hline
Full Test Set & 82,378 \\
\hline\hline
Action & 44,102 \\
Motion & 22,916 \\
Interaction &  15,360 \\
\hline
\hline
Cross Subject (CS) & 58,882 \\
Cross Action (CA) & 23,496 \\
\hline
\hline
Occlusion & 7,707 \\
Close-to-Geometry (C2G) & 1,727  \\
\hline
\end{tabular}
\end{center}
\caption{Numbers of frames in each test subset. We evaluate performance on
different subsets of the test data split by the scripted behavior
(Action/Motion/Interaction), subjects that were excluded from the training data
(cross-subject) and novel actions (cross-action). Finally, we evaluate on a
subset with significant occlusion (Occlusion) and a subset where many joints
were near scene geometry (Close-to-Geometry).}
\label{table:testset}
\end{table}

\section{Experiments}

\paragraph{Training data:} Our Geometric Pose Affordance (GPA) dataset has 304.8k images
of which 82k images are used for held-out test evaluation.  In addition, we use the 
MPII dataset \citep{mpii}, a large scale in-the-wild human pose dataset for training the 
2D pose module. It contains 25k training images and 2,957 validation images.
For the alternative baseline model (SIM), we use the MPII pre-trained ResNet \cite{xiao2018simple} 
to detect the 2D key points. We also evaluate performance when using the ground truth 
2D human pose, which serves as an upper-bound for the lifting-based method \cite{simple}. 

\paragraph{Implementation details:} We take a crop around the skeleton from the
original $1920 \times 1080$ image and isotropically resize to $256 \times
256$, so that projected skeletons have roughly the same size. Ground-truth target
2D joint location are adjusted accordingly. For ResNet-based method, following \cite{Zhou_2017_ICCV}, the
ground truth depth coordinates are normalized to $[0, 1]$.  The backbone for
all models is ResNet-50 \citep{resnet}. The 2D heat map/depth map
spatial resolution is $64 \times 64$ with one output channel per joint.
For test time 
evaluation, we scale each model prediction to match the average skeleton bone length
observed in the training. 

Models are
implemented in PyTorch with Adam as the optimizer. For the lifting-based method 
we use the same process as above to detect 2D joint locations and train the 
lifting network using normalized inputs and outputs by subtracting mean and 
dividing the variance for both 2D input and 3D ground-truth following \cite{simple}.

\paragraph{Evaluation metrics:} Following standard protocols defined in
\citep{mono_3dhp2017,h36m_pami}, we consider two evaluation metrics for
experiments: MPJPE (mean per-joint position error) and the 3DPCK (percent
correctly localized keypoints) with a distance threshold of 150 mm.  In
computing the evaluation metrics, root-joint-relative joint locations are
evaluated according to the each method original paper evaluation protocol.

\begin{figure}[t]
\centering
\includegraphics[width=0.3\linewidth]{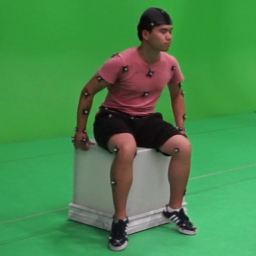}
\includegraphics[width=0.3\linewidth]{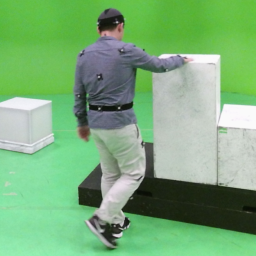}
\includegraphics[width=0.3\linewidth]{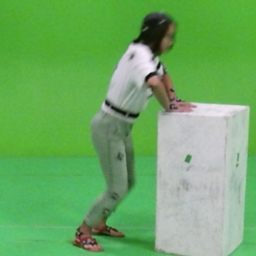}
\includegraphics[width=0.3\linewidth]{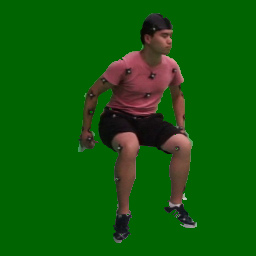}
\includegraphics[width=0.3\linewidth]{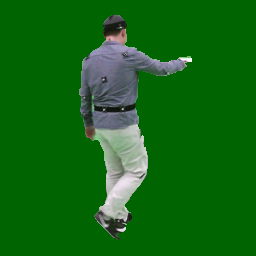}
\includegraphics[width=0.3\linewidth]{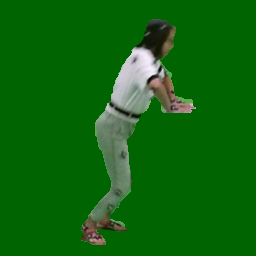}
\caption{We adopt Grabcut \cite{grabcut} and utilize the ground truth (joints, multi-layer depth, and markers) we have to segment subjects from background. If the joints and markers are occluded by the first-layer of multi-layer depth, we set them as background, otherwise they are set as foreground in grabcut algorithm. }
\label{fig:greenbackground}
\end{figure}

\paragraph{Evaluation subsets:} In addition to the three subsets --
Action, Motion, and Interaction -- that are inherited from the global split of the dataset
based on script contents, we also report test performance on 4 other subsets of the
test data: cross-subject (CS), cross-action (CA), occlusion, and close-to-geometry (C2G).
These are non-orthogonal splits of the test data which allow for finer characterizations 
of model performance and generalization in various scenarios:
\textbf{(1)} CS subset includes clips from held-out subjects to evaluate
generalization ability on unseen subjects and scenes; \textbf{(2)} CA subset includes
clips of held-out actions from same subjects from the training set;
\textbf{(3)} Occlusion subset includes frames with significant occlusions (at
least 10 out of 34 joints are occluded by objects); \textbf{(4)} Close-to-geometry
subset includes frames where subjects are close to objects (i.e. at least 8
joints have distance less than 175 mm to the nearest surface).

Statistics of these testing subsets are summarized in Table \ref{table:testset}.

\paragraph{Ablative study:} To demonstrate the contribution of each component, we
evaluate four variants of each model: the baseline models \textbf{ResNet / SIM-P / SIM-G} 
where \textbf{G} stands for ground-truth 2D joint input while \textbf{P} stands for predicted
2D joint input;
\textbf{ResNet-E / SIM-P-E / SIM-G-E / PoseNet-E}, models with encoded scene geometry
input; 
\textbf{ResNet-C / SIM-P-C / SIM-G-C / PoseNet-C}, the models with geometric consistency loss (GCL); \textbf{ResNet-F / SIM-P-F / SIM-G-F /PoseNet-F}, our full model
with both encoded geometry priors and GCL.

\begin{figure*}[t]
\begin{center}
   \includegraphics[width=1\linewidth]{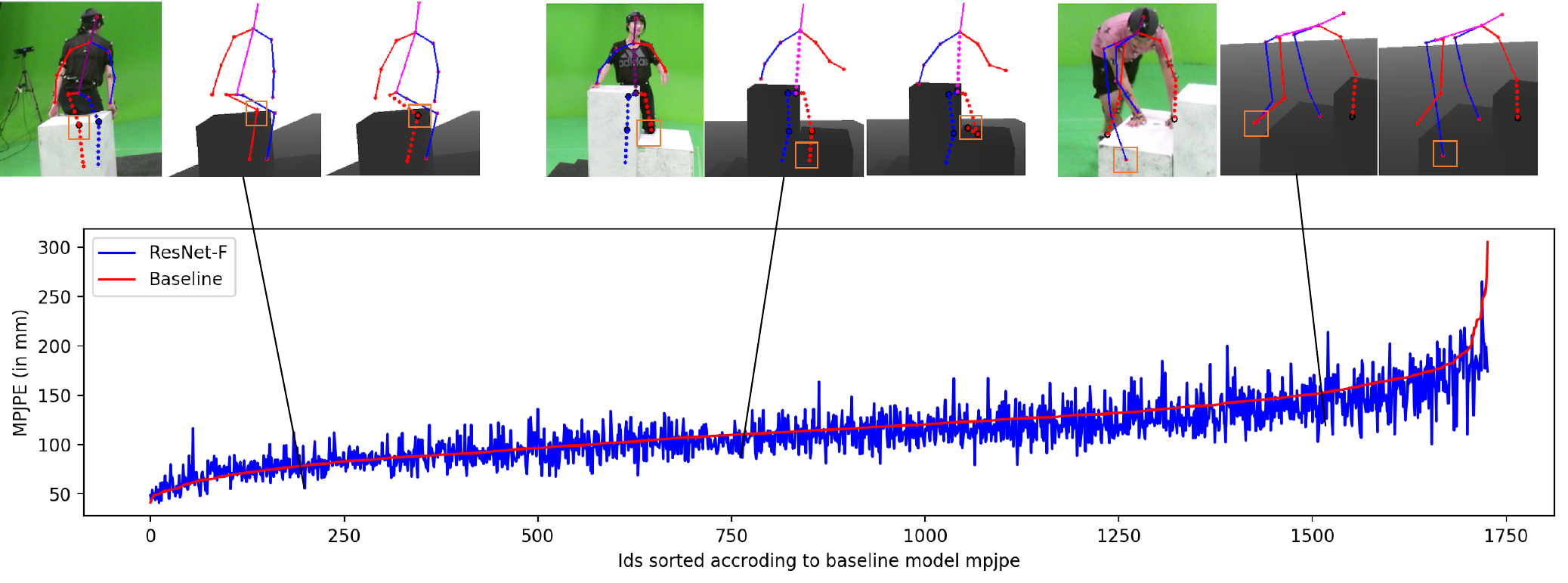}
\end{center}
   \caption{Distribution of prediction error (MPJPE) for ResNet-F and the baseline
   on the close-to-geometry test set. Examples are sorted in increasing order of 
   baseline MPJPE (red) with
   corresponding ResNet-F performances (GCL + encoding, in blue). We also highlight
   3 qualitative results, from left to right:  (a)
   case shows ResNet-F improve over the baseline with respect to the depth
   prediction. (b,c) cases show ResNet-F improves over the baseline in all
   $x,y,z$ axes. Furthermore, (b) demonstrates ResNet-F can even resolve 
   ambiguity under heavy occlusions with the aid of geometry information.  We
   show the image with the estimated 2D pose (after cropping), 1st layer of
   multi-layer depth map and whether the joint is occluded or not.
   \textbf{Legend:} hollow circles: occluded joints; solid dots: non-occluded
   joints; dotted lines: partially/completely occluded body parts; solid lines:
   non-occluded body parts.}
\label{fig:sucvis}
\end{figure*}

\begin{table}[t]
\begin{center}
\small
\begin{tabular}{l|c|c|c|c}
\hline
MPJPE & Baseline & ResNet-E & ResNet-C & ResNet-F \\
\hline
Full & 96.6 & 94.6 &  95.4 & 94.1 \\
\hline\hline
Action & 97.2 & 95.8 & 96.6 & 95.1 \\
Motion & 99.6 & 97.0 & 97.9 & 96.5 \\
Interaction & 89.7 & 87.5 & 88.3 & 87.4 \\
\hline
\hline
CS & 99.4 & 98.1 & 98.8 & 97.8 \\
CA & 89.2 & 85.8 &  86.7 & 85.6 \\
\hline
\hline
Occlusion & 120.5 & 116.1 & 117.9 & 115.1 \\
C2G & 118.1 & 113.2 & 116.3 & 111.5  \\
\hline
\end{tabular}
\end{center}
\caption{Prediction error (MPJPE) for ResNet-based models over the full test set as
well as different test subsets. Our proposed geometric encoding (ResNet-E) and geometric 
consistency loss (ResNet-C) each contribute to the performance of the full model 
(ResNet-F). Most significant reductions in error are for subsets involving significant 
interactions with scene geometry (Occlusion,C2G)}
\label{table:mpjpe-p-resnet}
\end{table}

\begin{table}[t]
\begin{center}
\small
\begin{tabular}{l|c|c|c|c}
\hline
MPJPE & Baseline & PoseNet-E & PoseNet-C & PoseNet-F \\
\hline
Full & 62.8 & 62.3 &  62.5 & 62.0 \\
C2G & 69.8 & 69.1 & 69.0 & 68.5  \\
\hline
Full & 78.8 & 78.5 &  78.2 & 78.1 \\
C2G & 91.9 & 91.4 & 91.6 & 89.4  \\
\hline
\end{tabular}
\end{center}
\caption{Prediction error (MPJPE) for ResNet-based models over the full test set as
well as different test subsets. Our proposed geometric encoding (PoseNet-E) and geometric 
consistency loss (PoseNet-C) each contribute to 
(PoseNet-F).}
\label{table:mpjpe-posenet}
\end{table}


\begin{table}[t]
\begin{center}
\small
\begin{tabular}{l|c|c}
\hline
Method & Full set & C2G \\
\hline
Lifting \cite{simple} & 91.2  & 112.8\\
ResNet-Baseline \cite{Zhou_2017_ICCV} & 96.6 & 118.1\\
PoseNet \cite{rootnet} & \underline{62.8} &  \underline{70.7}\\ 
I2L \cite{Moon_2020_ECCV_I2L-MeshNet} &  68.1 & 80.4 \\ 
DOPE \cite{dope} &  126.0 & 150.2\\
\hline
PoseNet (masked background)  & 64.4 & 78.7\\
\hline
Ours (PoseNet-F) & \textbf{62.0}  & \textbf{68.9}\\
\hline
\end{tabular}
\end{center}
\caption{We evaluated MPJPE (mm) for several recently proposed state-of-the-art architectures on our dataset. All models except DOPE were tuned on GPA training data.  We also trained and evaluated PoseNet on masked data (see Fig. 7) to limit implicit learning of scene constraints.}
\label{table:otherSOTA}
\end{table}

\begin{table}[t]
\begin{center}
\small
\begin{tabular}{l|c|c}
\hline
Dataset tested on /\ trained on  & GPA & H36M\\
\hline
H36M \cite{h36m_pami} & 118.8 & 61.4\\
GPA   & 62.8  & 110.9 \\
SURREAL \cite{varol17_surreal} & 126.2 & 142.4\\
3DPW \cite{inthewildeccv2018} & 125.5 & 132.5 \\
3DHP \cite{mono_3dhp2017} & 150.9 & 154.0\\
\hline
\end{tabular}
\end{center}
\caption{PoseNet models trained on our GPA dataset generalize well to other test datasets, outperforming models trained on H36M despite $\sim30\%$ fewer training examples \citep{crossdatasetevaluation}. We attribute this to the greater diversity of poses, occlusions and scene interactions present in GPA.}
\label{table:otherdataset}
\end{table}

\begin{table}[t]
\begin{center}
\small
\begin{tabular}{l|c|c|c|c}
\hline
PCK3D & Baseline & ResNet-E & ResNet-C & ResNet-F \\
\hline
Full & 81.9 & 82.5 &  82.3 & 82.9 \\
\hline\hline
Action & 81.4 & 81.8 & 81.6 & 82.0 \\
Motion & 80.7 & 81.5 & 81.6 & 82.0 \\
Interaction & 85.2 & 86.0 & 85.7 & 86.1 \\
\hline
\hline
CS & 81.3 & 81.7 & 81.5 & 82.0 \\
CA & 83.6 & 84.7 & 84.5 & 84.8 \\
\hline
\hline
Occlusion & 72.2 & 73.9 & 73.7 & 74.2 \\
C2G & 71.4 & 73.7 & 72.1 & 74.7  \\
\hline
\end{tabular}
\end{center}
\caption{Localization accuracy (PCK3D) follows similar trends to the mean errors 
reported in Table \ref{table:mpjpe-p-resnet}.}
\label{table:pck3d-p-resnet}
\end{table}

\subsection{Baselines}

To evaluate the difficulty of the GPA and provide context, we trained and 
evaluated a variety of recently proposed architectures for pose estimation including: 
DOPE \cite{dope},  Simple baseline \cite{simple}, ResNet-Baseline \cite{Zhou_2017_ICCV}, 
PoseNet \cite{rootnet}, 
and I2L \cite{Moon_2020_ECCV_I2L-MeshNet}. As data and code for training DOPE was not available,
we evaluated their released model. To account for systematic differences in the body joint definitions,
we utilized the average of hip joints as the DOPE coordinate origin (H36M-based models typically 
use the pelvis root joint as the origin). For the other architectures, we train and 
test on the GPA dataset following the original authors' hyperparameter settings. The results are illustrated 
in Table \ref{table:otherSOTA}. We can see a range of performance across different architectures,
ranging from 62.8 to 91.2 mm in MPJPE metric. Our full model built on the PoseNet architecture 
achieves the lowest estimation error. 

We break down the performance of the ResNet-based joint regression baseline on different 
subsets of data in Table \ref{table:mpjpe-p-resnet}.  We also list the corresponding PCK3D 
in Table \ref{table:pck3d-p-resnet}, which follows a similar pattern.
The motion, occlusion and close-to-geometry subsets prove to be the most challenging 
as they involve large numbers of frames where subjects interact with the scene geometry. 

\paragraph{Cross-dataset Generalization}

We find that pose estimators show a clear degree of over-fitting to the specific datasets on which they are trained on
\citep{crossdatasetevaluation}.  To directly verify whether the model trained on GPA generalizes to other datasets, we trained the high-performing PoseNet architecture using GPA and MPII \cite{mpii} data, and tested on several popular benchmarks: SURREAL~\cite{varol17_surreal}, 3DHP~\cite{mono_3dhp2017}, and 3DPW~\cite{inthewildeccv2018}. To evaluate consistently 
across test datasets, we only consider error on a subset of 14 joints which are common to all.  The MPJPE (mm) is 
illustrated in Table \ref{table:otherdataset}. We can see the model trained on GPA generalizes to other datasets with
similar or better generalization performance compared to the H36M trained variant. This is  surprising since H36M
train is roughly $30\%$ larger. We attribute this to the greater diversity of scene interactions, poses and occlusion patterns 
available in GPA train.

\subsection{Effectiveness of geometric affordance}

From Table \ref{table:mpjpe-p-resnet} we observe that incorporating geometric
as an input (ResNet-E) and penalizing predictions that violate constraints
during training (ResNet-C) both yield improved performance across all test
subsets. Not surprisingly, the full model (ResNet-F) which is trained to 
respect geometric context provided as an input achieves the best performance.
We can see from Table \ref{table:mpjpe-p-resnet} that the full model, ResNet-F decreases
the MPJPE by $2.1 mm$ over the whole test set. Among 4 subsets, the most
significant improvement comes on the occlusion and close-to-geometry subsets. Our
geometry-aware method decreases MPJPE in occlusion and C2G set by
5.4mm / 6.6mm and increase the PCK3D about 2\% / 3\%.  
Similar results hold for the SIM model.  The MPJPE is reduced when using either
the predicted (SIM-P-F) or ground-truth 2D joint locations (SIM-P-F) by 3mm
and 3.6mm respectively  (PCK3D improves 1.2\% and 1.1\%).  The improvement from 
SIM-G model is overall larger than SIM-P model due to the more accurate 2D 
location and better geometry information provided to the network.

\paragraph{Controlling for Visual Context} 
 One confounding factor in interpreting the power of geometric affordance for
 the ResNet-based model is 
 that while the baseline model doesn't use explicit geometric input, there is 
 a high degree of visual consistency between the RGB image and the underlying 
 scene geometry (e.g., floor is green, boxes are brighter white on top than 
 on vertical surfaces). As a result, the baseline model may well be implicitly
 learning some of the scene geometric constraints from images alone and 
 consequently decreasing the apparent size of the performance gap.
 
  To further understand whether the background pixels are useful or not for 3d pose estimation, we utilize Grabcut \cite{grabcut} to mask out background pixels. Specifically, we label the pixel belonging to markers, joints that are not occluded by the first-layer of multi-layer depth map as foreground, and occluded ones as background. Additionally, we dilate the skeleton constructed by all the joints and markers, use the inverse area as background area. We send these labels together with the image to OpenCV implementation Grabcut and get the foreground mask. We set the background color as green for better visualization as shown in Fig \ref{fig:greenbackground}. We use the model \cite{rootnet}, and train and test on the masked background images. We observe increased error on C2G from 70.7 mm to 78.7 mm MPJPE, which suggests that baseline models do take significant advantage of visual context in estimating pose. 
 
\paragraph{Errors by joint type:} We partition the 16 human joints into the limb
joints which are more likely to be interacting with scene geometry (out group)
and the torso and hips (in group).  The performance on these two subsets of
joints as well as individual joints is illustrated for the SIM model in 
Table \ref{table:eachjoint_mpjpe}. This verifies our assumption that limb 
joint estimation (wrist, elbow, knees, ankles) benefits more from incorporating 
geometric scene affordance. 

\begin{table}[t]
\begin{center}
\small
\begin{tabular}{l|c|c|c|c}
\hline
 MPJPE (mm) &  \multicolumn{2}{c}{Predicted Root} & \multicolumn{2}{c}{Ground Truth Root} \\
\hline
  & C2G & Full &   C2G & Full\\
\hline
\hline
 ResNet & 118.1 & 96.5 &    &   \\
 \hline
 ResNet-E & 116.0 & 95.4 &   113.2  & 94.6\\
 \hline
 ResNet-F & \textbf{115.1} & \textbf{94.7} & \textbf{111.5} & \textbf{94.1}\\
\hline
\hline
 SIM-P-B & 112.8 & 91.2 &   &  \\
 \hline
 SIM-P-E & 106.9 & 89.2 &  105.2  &  89.1\\
 \hline
 SIM-P-F & \textbf{105.1} & \textbf{88.9} &  \textbf{104.2}  &  \textbf{88.2}\\
\hline
\hline
 SIM-G-B & 79.8 & 68.2 & & \\
 \hline
 SIM-G-E & 76.3 & 65.3   & 74.2  &  64.8\\
 \hline
 SIM-G-F & \textbf{74.9} & \textbf{65.0} & \textbf{72.8}  &  \textbf{64.6}\\
\hline
\end{tabular}
\end{center}
\caption{The root joint depth is needed to offset the multi-layer depth map when encoding the scene geometry for relative pose estimation. Inaccurate root joint prediction limits but does not eliminate
the benefits of the geometric encoding.}
\label{table:mpjpe-j}
\end{table}

\paragraph{Error in predicted root joint:} 
Since our models predict joint depths relative to the root joint, it is necessary to offset the
multi-layer depth map values when encoding them as input. To make our evaluation more realistic, 
we also evaluated models using predicted root joint locations instead of using the ground-truth.
To estimate the (absolute) root joint depth, we utilize the model and training procedure
from \cite{rootnet} which estimates root joint depth based on the person bounding-box size
and image features.  This yields a mean root position error (MRPE) of 136.6mm with a mean
z-coordinate (depth) error of 116.8mm and x- and y-coordinate errors of 41.6mm and 35.2mm respectively.  Table \ref{table:mpjpe-j} shows the result of using this predicted root joint 
depth during encoding to offset the multi-depth map. Using predicted depth results in a 
loss of performance of about 1\% over the three methods (with the largest effect for ResNet)
but does not eliminate the benefits of geometric context.

\begin{table}[t]
\small
\begin{center}
\begin{tabular}{c|c|c||c|c}
\hline
MPJPE (mm)   & SIM-G & SIM-G-F & SIM-P & SIM-P-F \\
\hline
righthip & 17.1 & 15.5 & 22.5 & 20.7 \\
\hline
lefthip & 17.3 & 15.8 & 22.9 & 21.4  \\
\hline
spine1 & 48.3 & 44.9 & 63.5 & 62.4  \\
\hline
head & 55.1 & 51.7 & 70.4 & 69.0  \\
\hline
rightshoulder & 58.5 & 54.1 & 75.9 & 74.2   \\
\hline
leftshoulder & 61.0 & 56.7 & 78.7 & 75.4  \\
\hline
leftknee & 64.1 & 60.6 & 88.9 & 84.9  \\
\hline
rightknee & 64.6 & 61.3 & 91.8 & 87.2  \\
\hline
rightelbow & 81.4 & 75.1 & 108.5 & 103.6  \\
\hline
leftforeelbow & 84.5 & 81.8 & 104.1 & 102.9 \\
\hline
neck & 86.1 & 81.3 & 102.0 & 98.8  \\
\hline
rightankle & 86.6 & 83.2 & 127.5 & 122.1  \\
\hline
leftankle & 88.9 & 86.2 & 131.0 & 125.8  \\
\hline
rightwrist & 102.5 & 96.1 & 140.1 & 135.7  \\
\hline
leftwrist & 107.1 & 104.8 & 138.5 & 138.3  \\
\hline
\hline
in-group & 49.1 & 45.7 & 62.4 & 60.3  \\
\hline
out-group & 85.0 & 81.1 & 116.3 & 112.6  \\
\hline
\hline
all joints & 68.2 & 64.6 & 91.2 & 88.2  \\
\hline
\end{tabular}
\end{center}
\caption{Performance of the lifting network-based model \citep{simple} broken down by individual
joints and joint subsets. Baseline prediction error is higher for extremities (e.g., wrists and ankles)
which are inherently more difficult to localize.  These same joints typically show the largest
reduction in error from introducing geometric context.}
\label{table:eachjoint_mpjpe}
\end{table}

\paragraph{Computational Cost:} We report the average runtime over 10 randomly sampled images 
on a single 1080Ti in Table \ref{table:runningtime}. Timings for SIM do not include 2D keypoint
detection.  For comparison, we also include the run time for the PROX model of \cite{mpii_scene} 
which uses an optimization-based approach to perform geometry-aware pose estimation.  

\begin{figure*}[!bht]
\begin{center}
   \includegraphics[width=1\linewidth]{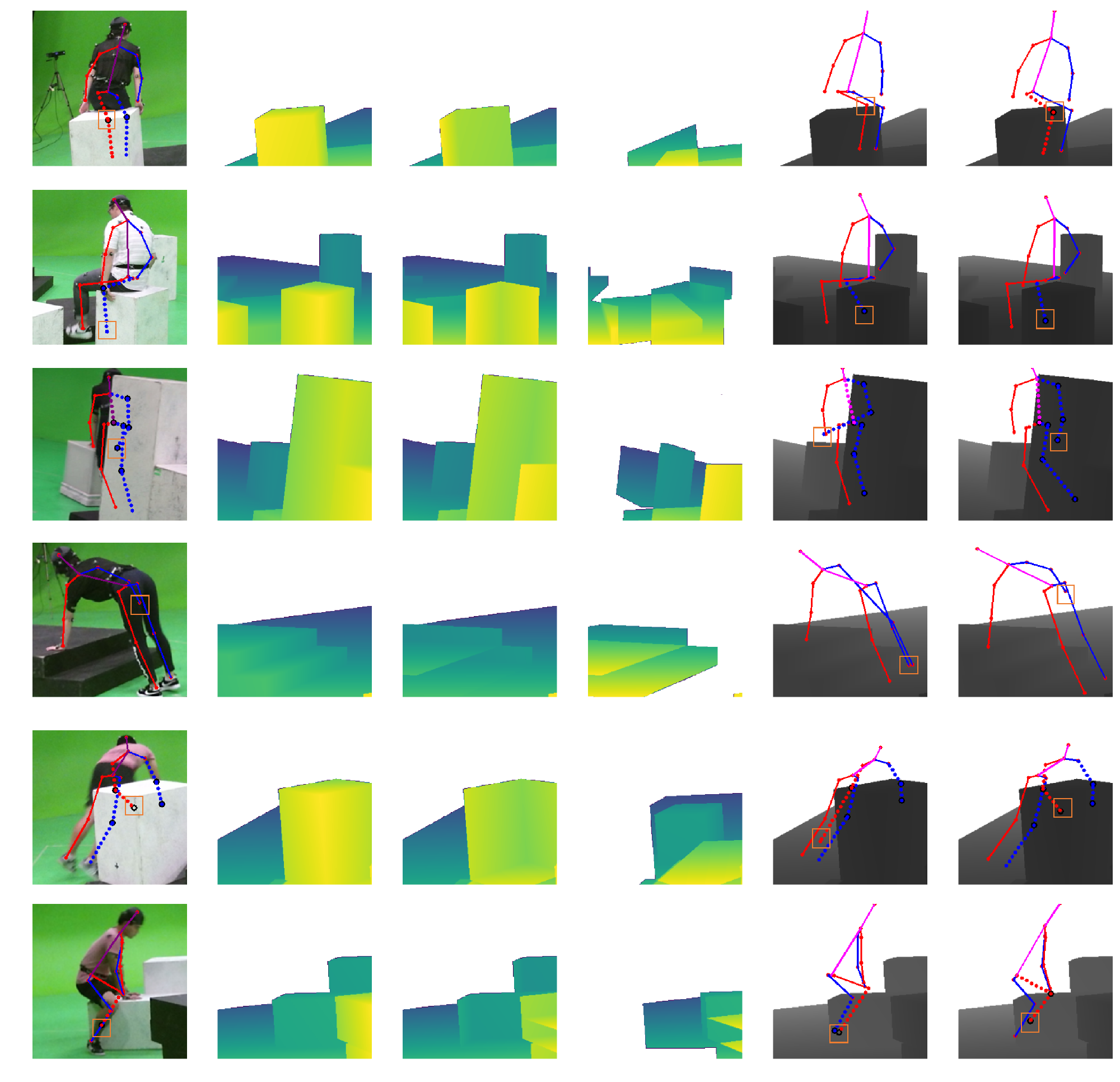}
\end{center}
   \caption{Visualization of the input images with the ground
   truth pose overlaid in the same view (blue and red indicate right and left sides
   respectively). Columns 2-4 depict the first 3 layers of
   multi-layer depth map. Column 5 is the baseline model prediction overlaid on the
   1st layer multi-layer depth map. Column 6 is the ResNet-F model prediction.
   The red rectangles highlight locations where the baseline model generates 
   pose predictions that violate scene geometry or are otherwise improved by
   incorporating geometric input. }
\label{fig:sucvis_example}
\end{figure*}

\paragraph{Qualitative results:} We show qualitative examples that high-light interaction
with geometry in Fig \ref{fig:sucvis} along with the distributions of the mean prediction 
error for the baseline and ResNet-F model over the close2geometry subset.  The geometry 
aware model is able to show most improvement for hard examples where the baseline error is large.
Further visualization of model predictions along with scene geometry encodings are shown in 
Fig \ref{fig:sucvis_example}. These examples demonstrate that ResNet-F has
better accuracy in both $xy$ localization and depth prediction and is often able to
resolve ambiguity under heavy occlusion where the baseline fails.

\begin{table}
\begin{center}
\small
\begin{tabular}{l|c}
\hline
 Method & Average Run Time \\
 \hline
 SIM \cite{simple} & 0.57 ms\\
\hline
 SIM-F & 0.64 ms\\
 \hline
 ResNet \cite{Zhou_2017_ICCV} & 0.29 s\\
\hline
 ResNet-F & 0.36 s\\
  \hline
 PROX \cite{mpii_scene} & 47.64 s\\
\hline
\end{tabular}
\end{center}
\caption{We compare the running time for our baseline backbone, our method, and another geometry-aware 3d pose estimation method PROX \cite{mpii_scene} averaged over 10 samples evaluated on a single GPU.}
\label{table:runningtime}
\end{table}

\section{Discussion and Conclusion}
 In this work, we introduce a large-scale dataset for exploring geometric 
pose affordance constraints. The dataset provides multi-view imagery with 
gold-standard 3D human pose and scene geometry, and features a rich variety 
of human-scene interactions. We propose using multi-layer depth
as a concise camera-relative representation for encoding scene geometry, and 
explore two effective ways to incorporate geometric constraints into training 
in an end-to-end fashion. There are, of course, many alternatives for representing 
geometric scene constraints which we have not yet explored. We
hope the availability of this dataset will inspire future work on
geometry-aware feature design and affordance learning for 3D human pose
estimation.

Broadly speaking, our techniques for encoding geometry yielded only modest reductions 
joint localization error ($\sim2-6\%$ depending on the base model).  We might have 
hoped for greater gains, but we expect that even the baseline models are implicitly 
learning something about scene constraints that are common across our dataset. 
Indeed, masking out the background yielded an $\sim11\%$ increase in baseline error.
There has been substantial success in training models that predict scene depth (2.5D) 
from monocular RGB inputs \citep{eigen,singleimagedepth} as well as full 3D representations such as
voxels \citep{ssn,shubhamfactrshapepose} or multilayer depth \citep{mdp}. This 
suggests that when geometric supervision is available, it may be useful to explore 
training systems that jointly estimate scene structure and 3D human pose in a 
multi-task setup.

In our experiments we focused on a setting where the scene geometric constraints were
available as input and highly accurate. While such prior knowledge is not available
in general (e.g., for a random photo on the web), we believe such data is readily 
accessible in many practical scenarios. The successful development of robust structure
from motion, SLAM, and specialized stereo or time-of-flight depth sensors makes geometric
scene information increasingly prevalent and easy to acquire. Assuming known camera and 
scene geometry as input appears practical in commercial applications where, e.g. robots 
navigate a well-mapped environment interacting with people or fixed cameras monitor 
human activity in a static workspace. We expect finding better techniques to incorporate 
such ``side information'' will offer a way to improve cross-scene/cross-dataset generalization 
and avoid some of the common over-fitting we currently observe when training and testing 
on individual datasets.

\section{Acknowledgments}
This project is supported by NSF grants IIS-1813785, IIS1618806, IIS-1253538,
CNS-1730158 and a hardware donation from NVIDIA. We thank reviewers for the 
valuable suggestions. We thank Shu
Kong and Minhaeng Lee for helpful discussion, John Crawford and Fabio Paolizzo
for providing support on the motion capture studio, and all the UCI friends who
contribute to collection of the dataset.





\bibliographystyle{model2-names}
\bibliography{refs}



\end{document}